\documentclass[final]{cvpr}

\usepackage{times}
\usepackage{epsfig}
\usepackage{graphicx}
\usepackage{amsmath}
\usepackage{amssymb}

\usepackage[pagebackref=true,breaklinks=true,colorlinks,bookmarks=false]{hyperref}
\usepackage{dblfloatfix}
\usepackage{balance}

\def\cvprPaperID{8601} 
\def\confYear{CVPR 2021}

\usepackage[utf8]{inputenc} 
\usepackage[T1]{fontenc}    
\usepackage{hyperref}       
\usepackage{url}            
\usepackage{booktabs}       
\usepackage{amsfonts}       
\usepackage{nicefrac}       
\usepackage{microtype}      

\usepackage{times}
\usepackage{epsfig}
\usepackage{graphicx}
\usepackage{amsmath}
\usepackage{amssymb}
\usepackage{bm}

\newcommand{\thetab}{\bm{\theta}}
\newcommand{\betab}{\bm{\beta}}

\newcommand{\bbeta}{\boldsymbol{\beta}}
\newcommand{\bdelta}{\boldsymbol{\delta}}
\newcommand{\btheta}{\boldsymbol{\theta}}

\newcommand{\sss}{\mathbf{s}}

\newcommand{\tb}{\mathbf{t}}

\newcommand{\mm}{\mathbf{m}}
\newcommand{\xx}{\mathbf{x}}

\newcommand{\rr}{\mathbf{r}}
\newcommand{\ttt}{\mathbf{t}}

\newcommand{\FF}{\mathbf{F}}
\newcommand{\KK}{\mathbf{K}}
\newcommand{\BB}{\mathbf{B}}

\usepackage{xspace}
\makeatletter
\DeclareRobustCommand\onedot{\futurelet\@let@token\@onedot}
\def\@onedot{\ifx\@let@token.\else.\null\fi\xspace}

\def\eg{\emph{e.g}\onedot} 
\def\ie{\emph{i.e}\onedot} 
 
\def\etc{\emph{etc}\onedot}

\makeatother

\title{Neural Descent for Visual 3D Human Pose and Shape}

\author{
Andrei Zanfir\quad Eduard Gabriel Bazavan\quad Mihai Zanfir \\ William T. Freeman \quad Rahul Sukthankar \quad Cristian Sminchisescu\\
\and
{\bf Google Research}\\
{\tt\small \{andreiz, egbazavan, mihaiz, wfreeman, sukthankar, sminchisescu\}@google.com}\\
}

\begin{document}

\maketitle

\begin{abstract}
We present deep neural network methodology to reconstruct the 3d pose and shape of people, including hand gestures and facial expression, given an input RGB image. We rely on a recently introduced, expressive full body statistical 3d human model, GHUM, trained end-to-end,
and learn to reconstruct its pose and shape state in a self-supervised regime. Central to our methodology, is a \emph{learning to learn and optimize} approach, referred to as HUman Neural Descent ({\bf HUND}), which avoids \emph{both} second-order differentiation when \emph{training the model parameters}, \emph{and} expensive \emph{state gradient descent} in order to accurately minimize a semantic differentiable rendering loss at test time. Instead, we rely on novel recurrent stages to update the pose and shape parameters such that not only losses are minimized effectively, but the process is meta-regularized in order to ensure end-progress.
{\bf HUND}'s symmetry between training and testing makes it the first 3d human sensing architecture to natively support different operating regimes including self-supervised ones. In diverse tests, we show that {\bf HUND} achieves very competitive results in datasets like H3.6M and 3DPW, as well as good quality 3d reconstructions for complex imagery collected in-the-wild.
\end{abstract}

\section{Introduction}

Automatic 3d human sensing from images and video would be a key, transformative enabler in areas as diverse as clothing virtual apparel try-on, fitness, personal well-being, health or rehabilitation, AR and VR for improved communication or collaboration, self-driving systems with emphasis to urban scenarios, special effects, human-computer interaction or gaming, among others. Applications in shopping, telepresence or fitness would increase human engagement and stimulate collaboration, communication, and the economy, during a lock-down.

The rapid progress in 3D human sensing has recently relied on volumetric statistical human body models \cite{SMPL2015,ghum2020} and supervised training. Most, if not all, state of the art architectures for predicting 2d, \eg, body keypoints \cite{cao2017realtime} or 3d, \eg, body joints, kinematic pose and shape \cite{dmhs_cvpr17,zanfir2018monocular,Kanazawa2018,kolotouros2019learning, sun2019human, doersch2019sim2real, humanMotionKanazawa19, kolotouros2019convolutional, arnab2019exploiting, xu2019denserac, kocabas2019vibe, varol18_bodynet, jackson20183d, pavlakos2017cvpr, yang20183d, zhou2017towards, rogez2016mocap, Fua17, mehta2017vnect,martinez17iccv, iskakov2019learnable} rely, \emph{ab initio}, at their learning core, on complete supervision. For 2d methods this primarily enters as keypoint or semantic segmentation annotations by humans, but for complex 3D articulated structures human annotation is both impractical and inaccurate. Hence for most methods, supervision comes in the form of synchronous 2d \emph{and} 3d ground truth, mostly available in motion capture datasets like Human3.6M \cite{Ionescu14pami} and more recently also 3DPW \cite{vonMarcard2018}. 

Supervision-types aside, the other key ingredient of any successful system is the interplay between 3d initialization using neural networks and non-linear optimization (refinement) based on losses computed over image primitives like keypoints, silhouettes, or body part semantic segmentation maps. No existing feedforward system, particularly a \emph{monocular} one, achieves \emph{both} plausible 3d reconstruction \emph{and} veridical image alignment\footnote{To be understood in the classical model-based vision sense of best fitting the model predictions to implicitly or explicitly-associated image primitives (or landmarks), within modeling accuracy.} without non-linear optimization -- a key component whose effectiveness for 3d pose estimation has been long since demonstrated \cite{sminchisescu_ijrr03,sminchisescu_cvpr03}.

The challenge faced by applying non-linear optimization in high-dimensional problems like 3d human pose and shape estimation stems from its complexity. On one hand, first-order model state updates are relatively inefficient for very ill-conditioned problems like \emph{monocular} 3d human pose estimation where Hessian condition numbers in the $10^{-3}$ are typical \cite{sminchisescu_ijrr03}. Consequently, many iterations are usually necessary for good results, even when BFGS approximations are used. On the other hand, nonlinear output state optimization is difficult to integrate as part of parameter learning, since correct back-propagation would require potentially complex, computationally expensive second-order updates, for the associated layers. Such considerations have inspired some authors \cite{kolotouros2019learning} to replace an otherwise desirable integrated learning process, with a dual system approach, where multiple non-linear optimization stages, supplying potentially improved 3d output state targets, are interleaved with classical supervised learning based on synchronized 2d and 3d data obtained by imputation. Such intuitive ideas have been shown to be effective practically, but remain expensive in training, and lack not just an explicit, integrated cost function, but also a consistent learning procedure to guarantee progress, in principle. Moreover, applying the system symmetrically, during testing, would still require potentially expensive non-linear optimization for precise image alignment. 

In this paper, we take a different approach and replace the non-linear gradient refinement stage at the end of a classical 3d predictive architecture with neural descent, in a model called HUND (Human Neural Descent). In HUND, recurrent neural network stages refine the state output (in this case the 3d human pose and shape of a statistical GHUM model \cite{ghum2020}) based on previous state estimates, loss values, and a context encoding of the input image, similarly in spirit to non-linear optimization. However, differently from models relying on gradient-based back-ends, HUND can be trained end-to-end using stochastic gradient descent, offers no asymmetry between training and testing, supports the possibility of potentially more complex, problem-dependent step updates compared to non-linear optimization, and is significantly faster. Moreover, by using such an architecture, symmetric in training and testing, with capability of refinement and self-consistency, we show, for the first time, that a 3d human pose and shape estimation system trained from monocular images can entirely bootstrap itself. The system would thus no longer necessarily require, the completely synchronous supervision, in the form of images and corresponding 3d ground truth configurations that has been previously unavoidable. Experiments in several datasets, ablation studies, and qualitative results in challenging imagery support and illustrate the main claims.


\noindent{\bf Related Work:} There is considerable prior work in 3d human modeling \cite{SMPL2015,ghum2020,dmhs_cvpr17,zanfir2018monocular,Rhodin_2018_ECCV,Kanazawa2018,kolotouros2019learning,ExPose:2020}, as well as the associated learning and optimization techniques \cite{sminchisescu_ijrr03,bogo2016}.
Systems combining either random 3d initialization or prediction from neural networks with non-linear optimization using losses expressed in terms of alignment to keypoints and body semantic segmentation masks exist \cite{bogo2016,zanfir2018monocular,kolotouros2019learning}. Black-box optimization has gained more interest in recent years \cite{andrychowicz2016learning, chen2017learning}, usually deployed in the context of meta-learning \cite{hospedales2020metalearning}. Our work is inspired in part by that of \cite{chen2017learning,hospedales2020metalearning} in which the authors introduce recurrent mechanisms to solve optimization problems, albeit in a different domain and for other representations than the ones considered in this work.
\cite{omran2018nbf} uses a neural network to directly regress the pose and shape parameters of a 3d body model from predicted body semantic segmentation. The network is trained in a mixed supervision regime, with either full supervision for the body model parameters or a weak supervision based on a 2d reprojection loss.
\cite{xiong2013supervised} propose to learn a series of linear regressors over SIFT \cite{lowe2004distinctive} features that produce descent directions analogous to an optimization algorithm for face alignment. Training is fully supervised based on 2d landmarks.
Similarly, \cite{trigeorgis2016mnemonic} learn a recurrent network, that given an input image of a face, iteratively refines face landmark predictions. The network is trained fully supervised and operates only in the 2d domain.
In \cite{tian2019regressing}, a cascade of linear regressors are learned to refine the 3d parameters of a 3d face model. Training is done over the entire dataset at a time (multiple persons with multiple associated face images) on synthetic data, in a simulated, mixed supervision regime.

\begin{figure*}[!ht]
\begin{center}
    \includegraphics[width=0.6\linewidth]{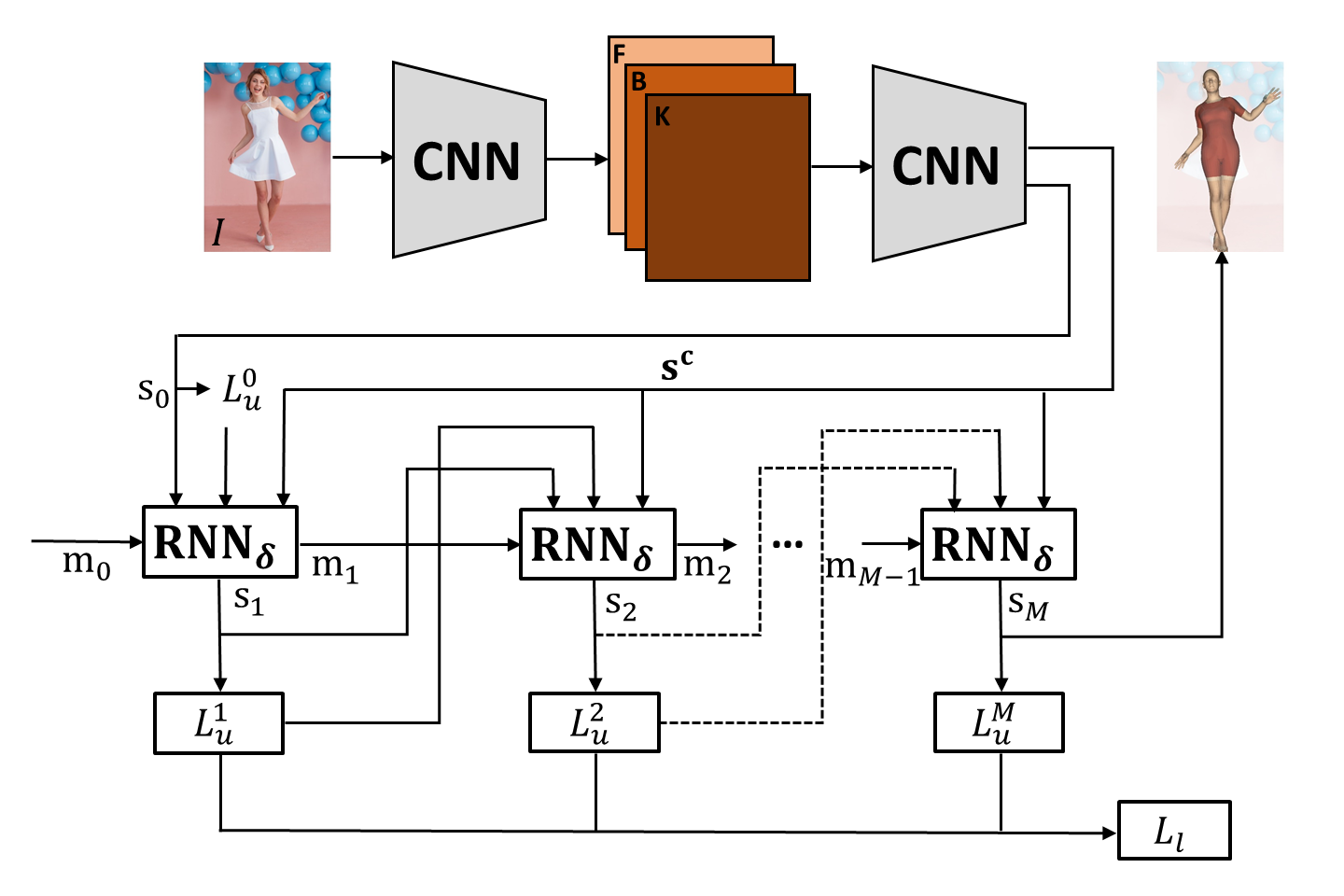}
\end{center}
\caption{\small Overview of our Human Neural Descent ({\bf HUND}) architecture for learning to estimate the state $\sss$ of a generative human model GHUM (including shape $\bbeta$ and pose $\btheta$, as well as person's global rotation $\rr$ and translation $\tb$) from monocular images. Given an input image, a first CNN extracts semantic feature maps for body keypoints ($\KK$) and part segmentation ($\BB$), as well as other features ($\FF$). These, in turn feed, into a second stage CNN that learns to compute a global context code $\sss^c$ as well as an initial estimate of the model state $\sss_0$. These estimates (and at later stages similar ones obtained recursively), together with the value of a semantic alignment loss $L_u$, expressed in terms of keypoint correspondences and differentiable rendering measures between model predictions and associated image structures, are fed into multiple refining RNN layers, with shared parameters $\bdelta$, and internal memory (hidden state) $\mm$. The alignment losses (which can be unsupervised, weakly-supervised or self-supervised, depending on available data) at multiple recurrent stages $M$ are aggregated into a learning loss $L_l$, optimized as part of the learning-to-learn process. The parameters are obtained using stochastic gradient descent, as typical in deep learning. The model produces refined state estimates $\sss$ with precise image alignment, but does not require additional gradient calculations for the recurrent stages neither in training  (\eg, second-order parameter updates), nor during testing (first-order state updates). It is also extremely efficient computationally compared to models relying on nonlinear state optimization at test time.}
\label{fig:pipeline}
\end{figure*}   

\begin{figure*}[!htbp]
\begin{center}
    \includegraphics[width=0.9\linewidth]{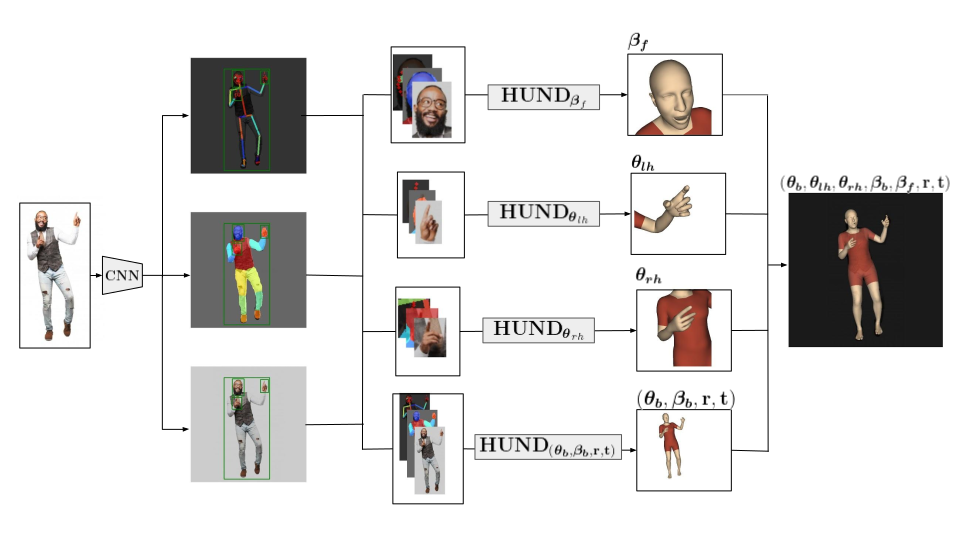}
\end{center}
\vspace{-10mm}
\caption{\small Our complete full body 3d sensing HUND network combines a feed-forward architecture to detect landmarks and semantically segment body parts with an attention mechanism that further processes the face, hands and the rest of the body as separate HUND predictive networks, with results fused in order to obtain the final, full body estimate. See fig.\ref{fig:pipeline} for the architecture of an individual HUND network.
}
\label{fig:hund-fullgp}
\end{figure*}

\section{Methodology}\label{sec:methodology}

We describe the elements of the proposed methodology, including the statistical 3D human body model GHUM, as well as the associated learning and reconstruction architecture used. We also cover the fusion of multiple architectures in order to obtain accurate estimates for the full body including hand gestures and facial expressions.

\subsection{Statistical 3D Human Body Model GHUM}
We use a recently introduced statistical 3d human body model called GHUM \cite{ghum2020}, to represent the pose and the shape of the human body. The model has been trained end-to-end, in a deep learning framework, using a large corpus of over 60,000 diverse human shapes, and $540,000$ human motions, consisting of 390,000 samples from CMU and 150,000 samples from Human3.6M (subjects S1, S5, S6, S7, S8). 
The model has generative body shape and facial expressions $\betab = \left( \boldsymbol{\beta_{b}, \betab_{f}} \right) $ represented using deep variational auto-encoders and generative pose $\thetab = \left( \boldsymbol{ \theta_{b}, \theta_{lh}, \theta_{rh}} \right)$ for the body, left and right hands respectively represented using normalizing flows \cite{zanfir2020weakly}. We assume a separable prior on the model pose and shape state $p(\thetab,  \betab)=p(\thetab) + p(\betab)$ where Gaussian components with $\mathbf{0}$ mean and unit $\mathbf{I}$ covariance, as typical in variational encoder and normalizing flow models. 
Given a monocular RGB image as input, our objective is to infer the pose $\thetab \in \mathbb{R}^{N_{p}\times1}$ and shape $\betab \in \mathbb{R}^{N_{s}\times1}$ state variables, where $N_{p}$ is the number of posing variables and $N_{s}$ is the length of the shape code, respectively. A posed mesh $\mathbf{M}(\thetab, \betab)$ has $N_{v}$ associated 3d vertices $\mathbf{V}=\{\mathbf{v}_{i}, i=1\ldots N_v\}$ with fixed topology given by the GHUM template. Because the rigid transformation of the model in camera space -- represented by a 6d rotation \cite{zhou2018continuity} $\rr\in\mathbb{R}^{6\times1}$ and a translation vector $\ttt\in\mathbb{R}^{3\times1}$ -- are important and require special handling, we will write them explicitly. The posed mesh thus writes $\mathbf{M}(\thetab, \betab, \rr, \ttt)$.

\paragraph{Camera model.} We assume a pinhole camera with intrinsics $C = [f_x, f_y, c_x, c_y]^\top$ and associated perspective projection operator $\xx_{2d} = \Pi(\xx_{3d}, C)$, where $\xx_{3d}$ is any 3d point. During training and testing, intrinsics \textit{for the full input image} are approximated, $f_x=\max(H, W), f_y=\max(H, W), c_x=W / 2, c_y=H / 2$, where $H, W$ are the input dimensions. Our method works with cropped bounding-boxes of humans, re-scaled to a fixed size 
of $480\times480$,
therefore we need to warp the image intrinsics $C$ into the corresponding crop intrinsics $C_c$
\begin{align}
    [C_c^\top 1]^\top = K [C^\top 1]^\top,
\end{align}
where $K \in \mathbb{R}^{5\times 5}$ is the scale and translation matrix, adapting the image intrinsics $C$. By using cropped intrinsics, we effectively solve for the state of the 3d model (including global scene translation) in the camera space of the input image. For multiple detections in the same image, the resulting 3d meshes are estimated relative to a common world coordinate system, into the \textit{same 3d scene}. At test time, when switching $C_c$ with $C$, the 3d model projection will also align with the corresponding person layout in the initial image.

\subsection{Learning Architecture}\label{sec:learn-arch}

The network takes as input a cropped human detection and resizes it to $480\times480$. A multi-stage sub-network produces features $\FF \in \mathbb{R}^{60\times60\times256}$, keypoint detection heatmaps $\KK \in \mathbb{R}^{60\times60\times137}$ and body-part segmentation maps $\BB \in \mathbb{R}^{60\times60\times15}$. These are embedded into a low-dimensional space, producing a code vector $\sss^c$ -- the superscript $c$ stands for context, \ie\ the optimization’s objective function context. We also append the cropped camera intrinsics $C_c$ to this context vector. At training time, a estimate $\sss_0$ of the initial GHUM state $\sss = [\thetab^\top, \betab^\top, \rr^\top, \ttt^\top]^\top$ is also produced.
To simulate model refinement\footnote{HMR\cite{Kanazawa2018} uses several  recursive output layers on top of a CNN prediction. However HMR does not use a formal RNN to recursively refine outputs based on a memory structure encompassing the previous estimates, the image reprojection (keypoint and semantic) error and the image feature code, as we do, which is the equivalent of a complete non-linear optimization context. Nor do we use a discriminator for pose as HMR, but instead rely on the kinematic normalizing flow prior of GHUM. Hence our approach is methodologically very different. See 
\S\ref{sec:exps} for quantitative evaluation. }, we employ a Recurrent Neural Network module $\mathbf{RNN}_{\bdelta}(\sss^c, \sss_i, \mm_i)$, where  $\mm_i$ is the memory (hidden state) at refinement stage $i$, and unroll the updates into $M$ stages (see fig.\ref{fig:pipeline})
\begin{equation}
\begin{bmatrix}
 \sss_i\\
\mm_i
\end{bmatrix} = \mathbf{RNN}_{\bdelta}(\sss_{i-1}, \mm_{i-1}, L_u^{i-1}, \sss^c).
\end{equation}
The loss at each stage $i$ is computed based on the labeling available at training time in the form of either 2d or 3d annotations. When both are missing, we are training with \textit{self-supervision}.  The self-supervised loss at each unit processing stage $i$ can be expressed as
\begin{align}\label{eq:ss-losses}
    L^i_{u}(\sss, \KK, \BB) = \lambda_k L_{k}(\sss_i, \KK) + \lambda_b L_{b}(\sss_i, \BB) + l(\thetab_i, \betab_i), 
\end{align}
where $l=-\log(p)$, $L_k$ is a 2d keypoint alignment loss, $L_b$ is a 2d semantic body part alignment (defined in terms of differentiable rendering), and $M$ is the total number of \textit{training} LSTM stages, while $\lambda_{k}$ and $\lambda_{b}$ are cross-validated scalar values which balance the loss terms.

The \textbf{keypoint alignment loss}, $L_k$, measures the reprojection error of the GHUM's model 3d joints w.r.t. the predicted 2d keypoints. The loss is defined as the 2d mean-per-joint position error (MPJPE)
\begin{align}
    L_{k}(\sss_t, \KK) = \frac{1}{N_j} \sum_i^{N_j} \|\mathbf{j}_{i}(\KK) - \Pi(\mathbf{J}_i(\sss_t), C_c)\|_2.
\end{align}
with $N_j$ keypoints, $\mathbf{j}_{i}(\KK)$ is the 2d location of the $i$-th 2d keypoint extracted from the the $\KK$ heatmap, and $\mathbf{J}_i(\sss_t)$ is the $i$-th 3d keypoint computed by posing the GHUM model at $\sss_t$.

The \textbf{body-part alignment loss}, $L_{b}$, uses the current prediction $\sss_t$ to create a body-part semantic segmentation image $I(\mathbf{M}(\sss_t), C_c)\in\mathbb{R}^{H\times W \times 15}$. Then we follow a soft differentiable rasterization process\cite{liu2019soft} to fuse probabilistic contributions of all predicted mesh triangles of the model, at its current state, with respect to the rendered pixels. In this way, gradients can flow to the occluded and far-range vertices.
To be able to aggregate occlusion states and semantic information, we append to each mesh vertex its semantic label, as a one-hot vector $\{0,1\}^{15\times 1}$, and a constant alpha value of $1$. The target body part semantic probability maps $\BB$ are also appended with a visibility value, equal to the foreground probability $\in [0,1]^{H\times W \times 1}$. The loss is the mean-per-pixel absolute value of the difference between the estimated and predicted semantic segmentation maps
\begin{align}
    L_{b}(\sss_t, \BB) = \frac{1}{HW} \sum_i^{HW} \|\BB_i - I(\mathbf{M}(\sss_t), C_c)_i\|_1.
\end{align}
For \textbf{body shape and pose}, we include two regularizers,  proportional to the negative log-likelihood of their associated Gaussian distributions
\begin{align}
    l(\thetab)=-\log p(\thetab) = \|\thetab\|_2^2, \;\;  l(\betab)=-\log p(\betab) = \|\betab\|_2^2.
\end{align}

When \textit{3d supervision} is available, we use the following unit training loss $L^{i}_{f}$, as well as, potentially, the other ones previously introduced 
in \eqref{eq:ss-losses}
for the self-supervised regime
\begin{align}
\begin{split}
    L^{i}_{f}(\sss) = \lambda_{m} L_{m}(\mathbf{M}(\sss_i), \widetilde{\mathbf{M}}) &+ \lambda_{3d} L_{3d}(\mathbf{J}(\sss_i), \widetilde{\mathbf{J}})
    \notag,
\end{split}
\end{align}
where $L_m$ represents the 3d vertex error between the ground-truth mesh $\widetilde{\mathbf{M}}$ and a predicted one, $\mathbf{M}(\sss_i)$--  obtained by posing the GHUM model using the predicted state $\sss_i$; $L_{3d}$ is the 3d MPJPE between the 3d joints recovered from the predicted GHUM parameters, $\mathbf{J}(\sss_i)$,  and the ground-truth 3d joints, $\widetilde{\mathbf{J}}$; $\lambda_{m}$ and $\lambda_{3d}$ are scalar values that balance the two terms.\\

For learning, we consider different losses $L_l$, including `sum', `last', `min' or `max', as follows
\begin{align}\label{eq:meta-losses}
    L_u^{\Sigma}(\sss,\KK, \BB)=\sum_{i=1}^{M} L_u^i(\sss_i,\KK, \BB) \nonumber\\
    L^{\rightarrow}_{u}(\sss, \KK, \BB) = L_u^M(\sss_M,\KK, \BB)\nonumber\\ 
    L_{u}^{\min}(\sss, \KK, \BB) = \min_{i=1}^{M} L_u^i(\sss_i,\KK, \BB) \nonumber\\
    L_{u}^{\max}(\sss, \KK, \BB) = \max_{i=1}^{M} L_u^i(\sss_i,\KK, \BB)
\end{align}
We also consider an \textit{observable improvement} (OI) loss for $L_l$ \cite{hospedales2020metalearning} 
\begin{align}\label{ml-oi}
    L_{u}^{oi} = \sum_{i=1}^{M} \min\{L_u^i - \min_{j<i}L_u^j, 0\}.
\end{align}

\noindent{\bf Multiple HUND networks for Body Pose, Shape, and Facial Expressions.} Capturing the main body pose but also hand gestures and facial expressions using a single network is challenging due to the very different scales of each region statistics. To improve robustness and flexibility we rely on 4 part networks, one specialized for facial expressions, two for the hands, and one for the rest of the body. Based on an initial person keypoint detection and semantic segmentation, we drive attention to face and hand regions as identified by landmarks and semantic maps, in order to process those features in more detail. This results in multiple HUND networks being trained, with estimates for the full body shape and pose fused in a subsequent step from parts (fig. \ref{fig:hund-fullgp}).

\section{Experiments}\label{sec:exps}

\noindent{\bf View of Experimental Protocols.} There is large variety of models and methods now available for 3d human sensing research, including body models like SMPL \cite{SMPL2015} and GHUM \cite{ghum2020}, or reconstruction methods like DMHS \cite{dmhs_cvpr17}, HMR \cite{Kanazawa2018}, SPIN \cite{kolotouros2019learning} \etc, set aside methods that combine random initialization or neural network prediction and non-linear refinement\cite{bogo2016,zanfir17}. To make things even more complex, some models are pre-trained on different 2d or 3d datasets and refined on others. A considerable part of this development has a historical trace, with models built on top of each-other and inheriting their structure and training sets, as available at different moments in time. 
Set that aside, multiple protocols are used for testing. For Human3.6M \cite{Ionescu14pami} only, there are at least 4: the ones originally proposed by the dataset creators, on the withheld test set of Human3.6M (or the representative subset Human80K \cite{Ionescu14}) as well as others, created by various authors, known as protocol 1 and 2 by re-partitioning the original training and validation sets for which ground truth is available. Out of these 2, only protocol 1 is sufficiently solid in the sense of providing a reasonably large and diverse test set for stable statistics (\eg, 110,000 images from different views in P1 vs. 13,700 in P2, from the same camera, at the same training set size of 312,000 configurations for both). Hence we use P1 for ablations and the official Human3.6M test set for more relevant comparisons. For some of the competing methods, e.g. SPIN\cite{kolotouros2019learning}, HMR\cite{Kanazawa2018} we ran the code from the released github repositories ourselves on the Human3.6M test set since numbers were not reported in the original publications. Results are presented in table \ref{tbl:H36MOfficial}. We will also use 3DPW \cite{vonMarcard2018} for similar reasons, or rather, in the absence of other options in the wild (30,150 training and 33,000 testing configurations). Testing all other model combinations would be both impractical and irrelevant, especially for new models like GHUM where most prior combinations are unavailable and impossible to replicate. As a matter of principle, 3D reconstruction models can be evaluated based on the amount of supervision received, be it 2d (for training landmark detectors {\bf \#2d det} or, additionally, for direct 3d learning {\bf \#2d}), {\bf \# 3d}, or synchronized {\bf \#2d-3d} annotations, the number of images used for self-supervision \#I, as well as perhaps number of parameters and run-time. In addition, ablations for each model, \eg, {\bf HUND}, would offer insights into different components and their relevance. We argue in support of this being one scientifically sound way of promoting diversity in the creation of new models and methods, rather than closing towards premature methodological convergence, weakly supported by unsustainable, ad-hoc, experimental combinatorics.
\begin{table*}[!htbp]
    \small
    \centering
    \begin{tabular}[t]{|l||r|r||r||r||r|r|r|r|}
    \hline
    \textbf{Method}  & {MPJPE-PA} & {MPJPE} & {MPJPE Trans} &  {\#2d det} & {\#2d} & {\#3d} & {\#2d-3d}&{\#I}\\ 
    \hline
    \hline
    \textbf{HMR (FS+WS) \cite{Kanazawa2018}} & $58.1$ & $88.0$& NR &129k & $111$k &$720$k  &$300$k  &$0$\\
    \hline
    \textbf{SPIN (FS+WS)} \cite{kolotouros2019learning} & NR & NR&NR&129k &$111$k & $720$k$/390$k & $300$k & $0$ \\
    \hline
    \textbf{HUND (FS+SS)} & $\mathbf{52.6}$ & $\mathbf{69.45}$& $152.6$ &80k&$0$ &$540$k & $150$k  & $54$k\\
    \hline
    \hline
    \textbf{HMR (WS) \cite{Kanazawa2018}} & $67.45$ & $106.84$&NR& 129k & 111k &$720$k &$0$ &$0$\\
    \hline    
    \textbf{HUND (SS)} & $\mathbf{66.0}$ & $\mathbf{91.8}$&$159.3$&80k &$0$ &$540$k &$0$ &$54$k\\
    \hline
    \end{tabular}
    \caption{\small Performance of different pose and shape estimation methods on the H3.6M dataset, with training/testing based on the representative protocol P1 (for self-supervised variants this only indicates the images used in testing). MPJPE-PA and MPJPE are expressed in mm. We also report the global translation of the body as this is supported by our fully perspective camera model (N.B. this is not supported by other methods which use an orthographic perspective model). We also compare different annotations used in the construction of different models, with a split into 2d (further differentiated into {\bf \#2d\,det} for training the joint landmarks and {\bf \#2d} for training the 3d learning algorithm), 3d and synchronized 2d-3d. The last column gives the number of images for self-supervised variants, \eg, HUND(SS), which do not use either 2d image keypoints or synchronized images and 3d mocap during training.}
\label{tbl:H36MP1}
\end{table*}

\begin{table}[!htbp]
    \small
    \centering
    \begin{tabular}[t]{|l||r|r||r||r||r|r|r|r|}
    \hline
    \textbf{Method}  & {MPJPE}\\ 
    \hline
    \hline
    \textbf{HMR (FS+WS) \cite{Kanazawa2018}} & $89$\\
    \hline
    \textbf{SPIN (FS+WS)} \cite{kolotouros2019learning} & $68$ \\
    \hline
    \textbf{HUND (FS+SS)} & $\mathbf{66}$ \\
    \hline
    \end{tabular}
    \caption{\small Results of different methods on the H3.6M official held-out test set. We achieve better results on a large test set of 900k images.}
\label{tbl:H36MOfficial}
\end{table}

\begin{table}[!htbp]
    \small
    \centering
    \begin{tabular}[t]{|l||r|r|}
    \hline
    \textbf{Method}  & {MPJPE-PA (mm)} & {MPJPE (mm)} \\ 
    \hline
    \hline
    \textbf{HMR (FS+WS) \cite{Kanazawa2018}} & $81.3$ & $130.0$ \\
    \hline
    \textbf{SPIN (FS+WS)\cite{kolotouros2019learning}} & $59.2$ & $96.9$ \\
    \hline
    \textbf{ExPose (FS+WS)\cite{ExPose:2020}} & $60.7$ & $93.4$ \\
    \hline
    \hline
    \textbf{HUND (SS)} & $63.5$ & $90.4$ \\
    \hline
    \textbf{HUND (FS+SS)} & $\mathbf{57.5}$ & $\mathbf{81.4}$\\
    \hline
    \end{tabular}
    
    \caption{\small Results on the 3DPW test set for different methods. Notice that a self-supervised version of HUND produces lower errors compared to the best supervised HMR implementation that includes not just synchronized $2d-3d$ training sets but also images with 2d annotation ground truth. A HUND model that includes asynchronous 2d-3d supervision, in addition to just unlabeled images, achieves the lowest error, and uses less training data than any other competitive method -- see also table \ref{tbl:H36MP1}.}
\label{tbl:3DPW}
\end{table}

For our \textbf{self-supervised (SS)} experiments, we employ two datasets containing images in-the-wild, COCO2017 \cite{lin2014microsoft} (30,000 images) and OpenImages \cite{OpenImages} (24,000), with no annotations in training and testing. We refer to \textbf{weakly-supervised (WS)} experiments as those where ground truth annotations are available, \eg\ human body keypoints. We do not rely on these but some other techniques including HMR and SPIN do, hence we make this distinction in order to correctly reflect their supervision level.

For \textbf{fully supervised (FS)} experiments, we employ H3.6M and 3DPW. Because we work with the newly released GHUM model, we retarget the mocap raw marker data from H3.6M to obtain accurate 3d mesh supervision for our model \cite{ghum2020}. Because the ground-truth of 3DPW is provided as SMPL 3d meshes, we fit the GHUM model by using an objective function minimizing vertex-to-vertex distances between the two corresponding meshes. 

\noindent{\bf Architecture Implementation.} To predict single-person keypoints and body part segmentation, we train a multi-task network with ResNet50 \cite{he2016identity} backbone (the first CNN in our pipeline, see fig.\ref{fig:pipeline} and fig.\ref{fig:hund-fullgp})\cite{dmhs_cvpr17}.
We have $137$ 2d keypoints as in \cite{cao2019openpose} and $15$ body part labels, as in \cite{Guler2018DensePose}. This network has $34$M trainable parameters.
The self-supervised training protocol for \textbf{HUND} assumes only images are available and we predict 2d body keypoints and body part labels during training, in addition to body shape and pose regularizers. For the embedding model (the second CNN in the pipeline) predicting $\sss^c$, we use a series of $6$ convolutional layers with pooling, followed by a fully connected layer. We use $M = 5$ LSTM \cite{hochreiter1997long} stages as RNNs for \textbf{HUND} and we set the number of units to $256$, which translates to $525$k parameters. In total there are $950$k trainable parameters for the 3D reconstruction model. We train with a batch size of $32$ and a learning rate of $10^{-4}$ for $50$ epochs. For experiments where we train HUND using {\bf FS+SS}, we use a mixed schedule, alternating between self-supervised and fully-supervised batches. Training takes about 72 hours on a single Nvidia Tesla P100 GPU. The runtime of our prediction network for a single image is $0.035$s and $0.02$s for \textbf{HUND}, on an Nvidia RTX 2080 GPU.  

\begin{figure*}[!htbp]
\begin{center}
    \includegraphics[width=0.95\linewidth]{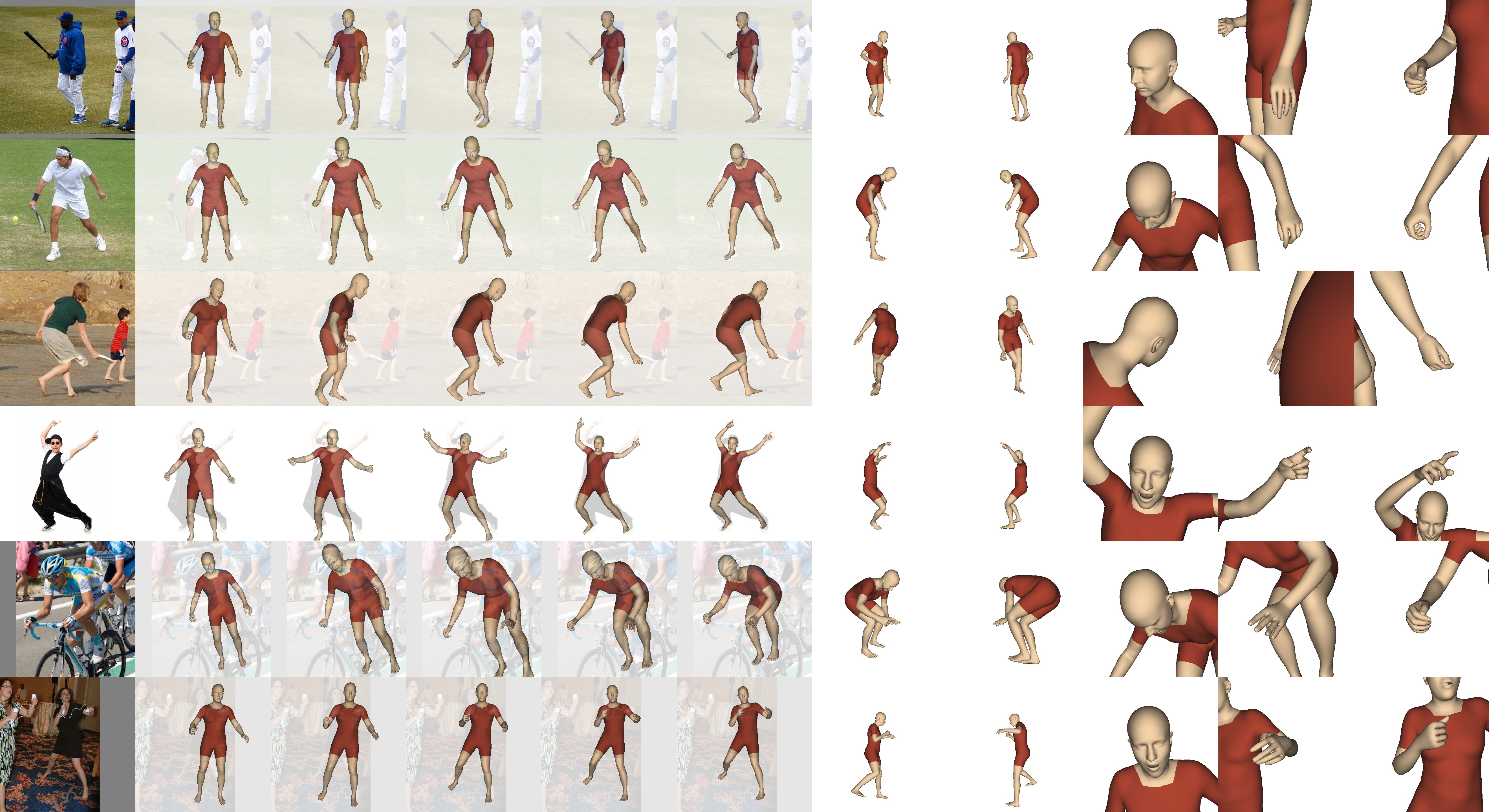}
\end{center}
\vspace{-4mm}
\caption{\small Visual 3d pose and shape configurations of GHUM sampled by {\bf HUND} during optimization. First column shows the input image, columns 2-6 illustrate GHUM estimates at each HUND stage. Columns 7 and 8 show visualizations of the GHUM state from different viewpoints, after HUND terminates. Columns 9, 10 and 11 show close up views for the reconstructed face expressions, left and right hands.
}
\label{fig:hund-optim-visuals}
\end{figure*}

\noindent{\bf Evaluation and discussion.} Multiple experiments are run for different regimes. Quantitative results are presented in tables \ref{tbl:H36MP1}, \ref{tbl:H36MOfficial} and \ref{tbl:3DPW} for Human3.6M and 3DPW respectively.
\begin{figure*}[!htbp]
\vspace{-4mm}
\begin{center}
    \includegraphics[width=0.4\linewidth]{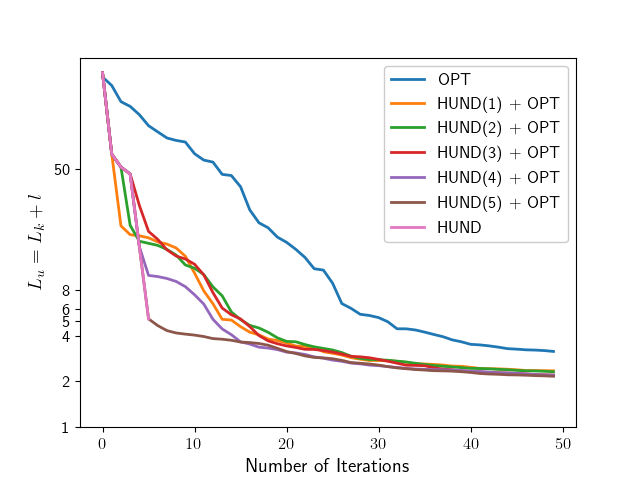}~~
    \includegraphics[width=0.4\linewidth]{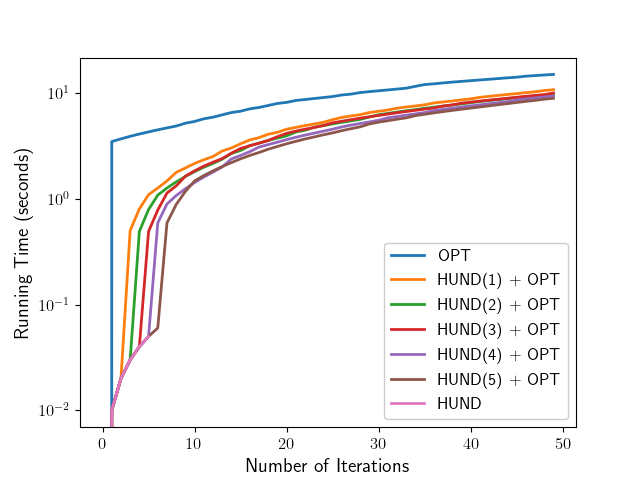}
\end{center}
\vspace{-3mm}
\caption{Behavior of different optimization methods including standard non-linear gradient-based BFGS, HUND(5), as well as variants of HUND$(i), i \leq 5$, initializing BFGS, in order to assess progress and the quality of solutions obtained along the way (left). Corresponding cumulative run-times are shown on the right. Observe that HUND produces a good quality solution orders of magnitude faster than gradient descent (note log-scales on both plots). End refinement using gradient descent improves results, although we do not recommend a hybrid approach--- here we only show different hybrids for insight. This shows one optimization trace for a model initialized in A-pose and estimated given one image from Human3.6M, but such behavior is typical of aggregates, see \eg, fig. \ref{fig:averaged_optimization_and_runtime}. See also fig. \ref{fig:hund-optim-visuals} for visual illustrations of different configurations sampled by HUND during optimization.}
\label{fig:single_image_optimization_and_runtime}
\end{figure*}
A detailed analysis of optimization behavior for one image is given in fig. \ref{fig:single_image_optimization_and_runtime} as well as, in aggregate, in fig. \ref{fig:averaged_optimization_and_runtime}. Visual reconstructions at different HUND optimization stages, for several images, are given in fig. \ref{fig:hund-optim-visuals}. 
\begin{figure*}[!htbp]
\vspace{-4mm}
\begin{center}
    \includegraphics[width=0.4\linewidth]{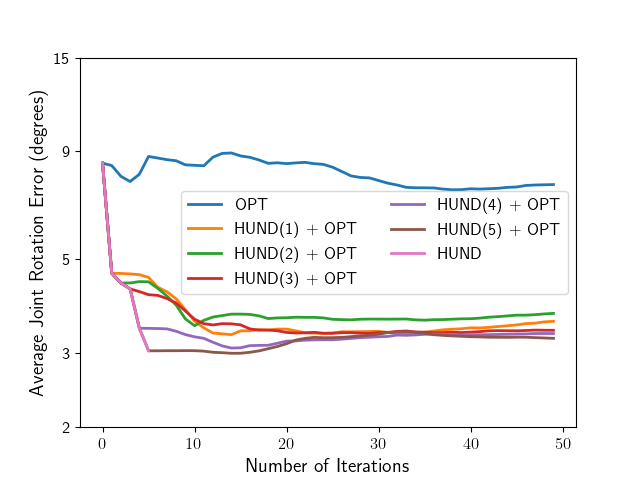}~~
    \includegraphics[width=0.4\linewidth]{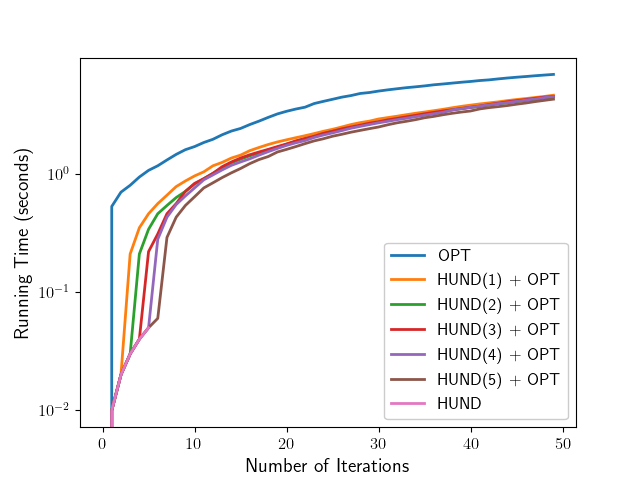}    
\end{center}
\vspace{-3mm}
\caption{\small Optimization statistics for different methods, aggregated over 100 different poses (estimation runs) from Human3.6M. We initialize in an A-pose and perform monocular 3d pose and shape reconstruction for GHUM under a HUND (FS+SS) model, as well as non-linear optimization baselines. On the left we show per-joint angle averages w.r.t. ground truth. On the right we show running times in aggregate for different types of optimization. One can see that BFGS descent under a keypoint+prior loss tends to be prone to inferior local optima compared to different HUND hybrids, which on average find significantly better solutions. The plot needs to be interpreted in proper context, as aggregates meant to show distance and run-time statistics per iteration. Hence, they may not be entirely representative of any single run, but for a singleton see \eg, fig. \ref{fig:single_image_optimization_and_runtime}.
}
\label{fig:averaged_optimization_and_runtime}
\end{figure*}

\begin{figure*}[!htbp]
\begin{center}
    \includegraphics[width=0.95\linewidth]{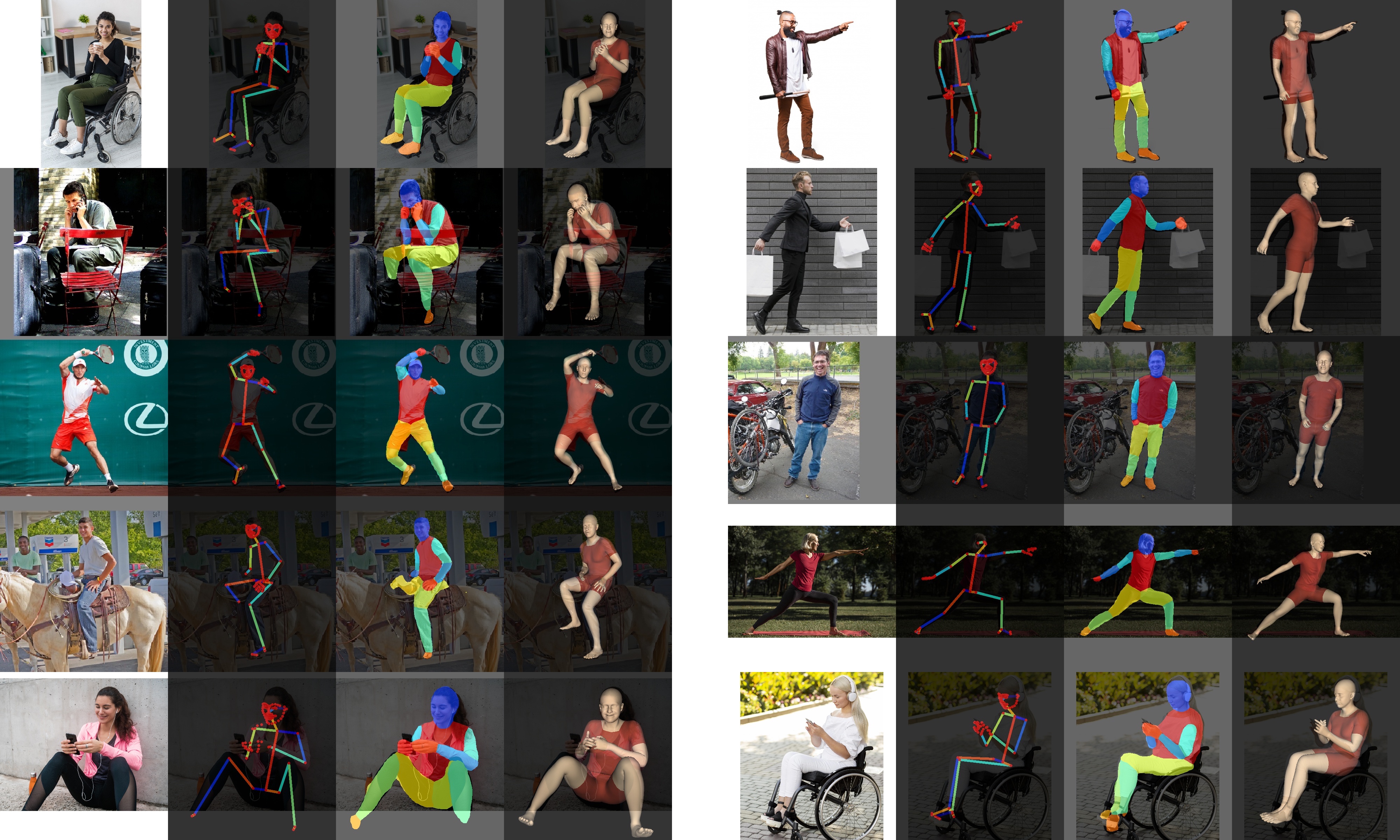}
\end{center}
\vspace{-4mm}
\caption{\small Visual 3d reconstruction results obtained by {\bf HUND}. Given initial 2d predictions for body, face and hand keypoints, and initial predictions for semantic body part labelling, the neural descent network predicts the 3d GHUM pose and shape parameters. Best seen in color. For other examples and videos see our Sup. Mat. 
}
\label{fig:visuals1}
\end{figure*}

\begin{table}[!htbp]
    \small
    \centering
    \begin{tabular}[t]{|l||r|r|}
    \hline
    \textbf{Loss}  & {MPJPE-PA (mm)} & {MPJPE (mm)}\\ 
    \hline
    \hline
    $L_f^\rightarrow$ & $\mathbf{58.50}$ & $\mathbf{80.16}$\\
    \hline
    $L_f^{\Sigma}$ & $59.91$ & $83.26$ \\
    \hline
    $L_f^{\min}$ & $78.61$ & $122.60$\\
    \hline
    $L_f^{oi}$\ & $79.35$ & $123.90$\\
    \hline
    $L_f^{\max}$ & $83.80$ & $128.0$\\
    \hline
    \end{tabular}
    \caption{\small Impact assessment of different meta-losses used in HUND (FS), trained on the Human3.6M dataset, following protocol 1. The last and sum losses perform similarly well, with others following at a distance.}
\label{tbl:losses_ablation}
\end{table}
We also study the impact of different meta-learning losses, as given in \eqref{eq:meta-losses} and \eqref{ml-oi}, on the quality of results of HUND. We use a HUND (FS) model trained and evaluated on Human3.6M (protocol 1). From table \ref{tbl:losses_ablation} we observe that the last ($L_f^\rightarrow$) and sum ($L_f^{\Sigma}$) losses perform best, whereas others produce considerably less competitive results, by some margin, for this problem.
Finally, we show qualitative visual 3d reconstruction results, from several viewpoints, for a variety of difficult poses and backgrounds in fig. \ref{fig:visuals1}. \emph{Please see our Sup. Mat. for videos!}

\noindent{\bf Ethical Considerations.} Our methodology aims to decrease bias by introducing flexible forms of self-supervision which would allow, in principle, for system bootstrapping and adaptation to new domains and fair, diverse subject distributions, for which labeled data may be difficult or impossible to collect upfront. Applications like visual surveillance and person identification would not be effectively supported currently, given that model's output does not provide sufficient detail for these purposes. This is equally true of the creation of potentially adversely-impacting deepfakes, as we do not include an appearance model or a joint audio-visual model.

\section{Conclusions}

We have presented a neural model, {\bf HUND}, to reconstruct the 3d pose and shape of people, including hand gestures and facial expressions, from image data. In doing so, we rely on an expressive full body statistical 3d human model, GHUM, to capture typical human shape and motion regularities. Even so, accurate reconstruction and continuous learning are challenging because large-scale diverse 3d supervision is difficult to acquire for people, and because the most efficient inference is typically based on non-linear image fitting. This is however difficult to correctly `supra'-differentiate, to second order, in training and expensive in testing. To address such challenges, we rely on self-supervision based on differentiable rendering within \emph{learning-to-learn} approaches based on recurrent networks, which avoid expensive gradient descent in testing, yet provide a surrogate for robust loss minimization. 
{\bf HUND} is tested and achieves very competitive results for datasets like H3.6M and 3DPW, as well as for complex poses, collected in challenging outdoor conditions. HUND's learning-to-learn and optimize capabilities, and symmetry between training and testing, can make it the first architecture to demonstrate the possibility of bootstraping a plausible 3d human reconstruction model without initial, synchronous (2d, 3d) supervision.

\clearpage
\balance
{\small
\bibliographystyle{ieee_fullname}
\bibliography{egbib}
}
\end{document}